\documentclass[runningheads]{llncs}
\usepackage{graphicx}

\usepackage{tikz}
\usepackage{comment}
\usepackage{amsmath,amssymb} %
\usepackage{color}

\usepackage[accsupp]{axessibility}  %

\usepackage{hyperref}

\makeatletter
\newcommand{\printfnsymbol}[1]{%
  \textsuperscript{\@fnsymbol{#1}}%
}
\makeatother

\usepackage{xspace}
\usepackage{graphicx}
\usepackage{amsmath}
\usepackage{amssymb}
\usepackage{booktabs}
\usepackage{subcaption}

\usepackage{pifont}%
\newcommand{\cmark}{\ding{51}}%

\newcommand*{\ie}{i.e.\@\xspace}
\newcommand*{\eg}{e.g.\@\xspace}

\newcommand*{\etal}{et al.\@\xspace}
\newcommand{\Expect}{{\rm I\kern-.3em E}}

\begin{document}
\pagestyle{headings}
\mainmatter
\def\ECCVSubNumber{5265}  %

\title{Exploring Fine-Grained Audiovisual Categorization with the SSW60 Dataset} %

\titlerunning{Exploring Fine-Grained Audiovisual Categorization}

\author{
{Grant Van Horn\index{Van Horn, Grant}\inst{1}\thanks{~The first two authors contributed equally. \url{https://github.com/visipedia/ssw60}} \quad Rui Qian\inst{1}\printfnsymbol{1} \quad Kimberly Wilber\inst{2}} \\
{Hartwig Adam\inst{2} \quad Oisin Mac Aodha\index{Mac Aodha, Oisin}\inst{3} \quad Serge Belongie\inst{4}} \\ %
}

\authorrunning{Van Horn et al.}
\institute{$^1$Cornell University \quad $^2$Google \\
$^{3}$University of Edinburgh \quad $^4$University of Copenhagen }

\maketitle

\begin{abstract}

We present a new benchmark dataset, Sapsucker Woods 60 (SSW60), for advancing research on audiovisual fine-grained categorization. 
While our community has made great strides in fine-grained visual categorization on images, the counterparts in audio and video fine-grained categorization are relatively unexplored.
To encourage advancements in this space, we have carefully constructed the SSW60 dataset to enable researchers to experiment with classifying the same set of categories in three different modalities: images, audio, and video. 
The dataset covers 60 species of birds and is comprised of images from existing datasets, and brand new, expert curated audio and video datasets. 
We thoroughly benchmark audiovisual classification performance and modality fusion experiments through the use of state-of-the-art transformer methods. 
Our findings show that performance of audiovisual fusion methods is better than using exclusively image or audio based methods for the task of video classification. 
We also present interesting modality transfer experiments, enabled by the unique construction of SSW60 to encompass three different modalities. 
We hope the SSW60 dataset and accompanying baselines spur research in this fascinating area.

\end{abstract}

\section{Introduction}
\label{sec:intro}

Image-based fine-grained visual categorization (FGVC) of natural world categories has seen impressive performance gains over the last decade of research. This progression has been fueled by both larger datasets and improved techniques for classification. For example, consider the domain of bird species classification. The popular CUB200~\cite{wah2011caltech} dataset (covering 200 classes of birds, each with 30 train and 30 test images) has seen top-1 accuracy improve from $10.3$\%~\cite{wah2011caltech} to over $91.7$\%~\cite{he2021transfg}. This dataset motivated the construction of the larger and better curated NABirds~\cite{van2015building} dataset (covering 400 species of birds, each with 60 train and 60 test images), which subsequently gave rise to the larger iNaturalist competition datasets~\cite{iNat2021Comp}. 
The latest dataset in this series has 1,486 species of birds, most with 300 training examples, and the winners of the 2021 iNaturalist competition~\cite{iNat2021Comp} achieved 94\% top-1 accuracy on these species (using geographic location information). 
The release of the CUB200 dataset was a catalyst for FGVC research, motivating the construction of improved datasets as well as providing the means to benchmark progress. 
But what about the challenge of fine-grained categorization (FGC) in modalities besides images?

\begin{figure}[t]%
\centering
\includegraphics[width=0.9\linewidth]{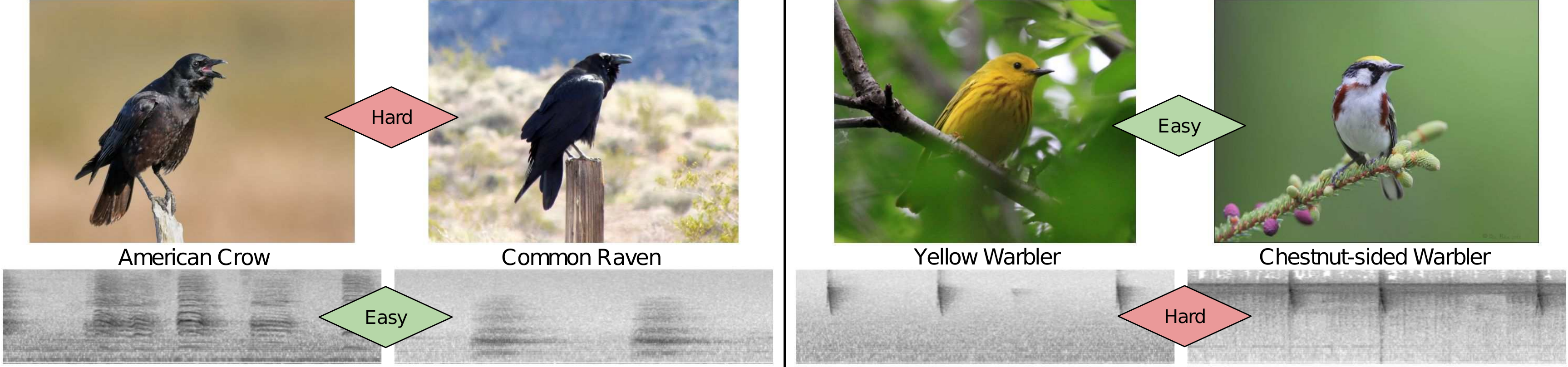}
\caption{
\textbf{Why audiovisual?}
Left: American Crow  and Common Raven  are visually confusing but aurally distinguishable, as illustrated by the spectrograms. 
Right: Yellow Warbler  and Chestnut-sided Warbler are aurally confusing but visually distinguishable. 
Individually, audio and visual modalities have both advantages and disadvantages.
We present the Sapsucker Woods 60 dataset (SSW60), a new dataset to facilitate work in fine-grained audiovisual categorization. 
}
\label{fig:audiovisual_challenges}
\end{figure}

Audio and video modalities receive less attention than images for the task of fine-grained categorization. 
What opportunities and challenges do these modalities present (see Fig.~\ref{fig:audiovisual_challenges})? 
More importantly, which existing datasets allow us to study cross-modality performance %
and where do they fall short?
Large-scale audiovisual datasets such as AudioSet~\cite{gemmeke2017audio} and VGGSound~\cite{chen2020vggsound}, provide a class hierarchy more akin to coarse grained categories as perceived by humans than fine-grained categories as typically used in the context of FGC (with classes such as ``Chirp, tweet'' and ``Hoot'' for bird vocalizations in AudioSet). 

There are a few existing bird video datasets~\cite{ge2016exploiting,saito2016ibc127,zhu2018fine}, each focused primarily on benchmarking the performance of video frame classification, as opposed to cross-modality or audiovisual analysis. 
The YouTube-Birds dataset~\cite{zhu2018fine} almost checks all the boxes,
except upon close inspection we find multiple inconveniences, \eg it consists of a collection of YouTube links, of which at least 7\% are broken at the time of writing, it contains labelling errors (typical for fine-grained datasets curated by non-experts), and the videos are not trimmed to the content of interest and are thus long and unwieldy.  
Both VB100~\cite{ge2016exploiting} and IBC127~\cite{saito2016ibc127} sampled their videos from higher quality data sources, but they each lack the full complement of unpaired audio and image modalities that we require for exploring audiovisual categorization. Finally, none of the prior art show the utility of audiovisual fusion methods for FGC. 

In this paper, we aim to fill this dataset gap and open up new avenues of research in FGC. 
Our new dataset, SSW60, spans 60 species of birds that all occur in a specific geographic location: Sapsucker Woods in Ithaca, New York, (unlike the random collection of species present in the existing video datasets~\cite{ge2016exploiting,saito2016ibc127,zhu2018fine}). SSW60 contains a new collection of expert curated ten-second video clips for each species, totaling 5,400 video clips. 
SSW60 also contains an ``unpaired'' expert curated set of ten-second audio recordings for the same set of species, totaling 3,861 audio recordings. 
Finally, we also collate image data for the same species from the existing  expert curated NABirds dataset~\cite{van2015building} and the  citizen science collected iNat2021 dataset~\cite{van2021benchmarking}. 

With this new dataset in hand, we perform a thorough investigation of audiovisual classification performance. 
Our baseline methods utilize state-of-the-art backbones trained on visual and audio modalities. 
We experiment with several different fusion methods to combine information from both modalities and make audiovisual informed classifications. 
These experiments reveal that audiovisual methods outperform their respective single modality counterparts, advancing the state of the art for fine-grained bird species classification. 
As SSW60 contains images and unpaired audio examples, we conduct additional experiments to investigate the utility of pretraining on these individual modalities prior to working with video.
We identify several insights from these experiments, including the unexpected negative impact of pretraining on high quality images, and the high utility of pretraining on unpaired audio samples.

In summary, we make the following contributions: 
1) A new fine-grained dataset that contains expert curated video and audio data for a shared set of object categories.  
2) A detailed analysis of cross-modality learning in the context of fine-grained object categories, as well as benchmark results for fine-grained audiovisual categorization.

\section{Related Work}
\label{sec:rel_work}

\subsection{Image, Audio, and Video Datasets}

\noindent{\bf Fine-Grained Image Datasets.} The most commonly used classification datasets in computer vision predominantly deal with coarse-grained object reasoning, \eg~\cite{russakovsky2015imagenet,zhou2017places,OpenImages2,everingham2010pascal,lin2014microsoft,gupta2019lvis}.   
In contrast, fine-grained datasets contain subordinate categories that can be much more challenging for non-expert human annotators to discriminate. 
There are many fine-grained datasets spanning a wide range of visual concepts including airplanes~\cite{maji2013fine,vedaldi2014understanding}, automobiles~\cite{krause20133d,lin2014jointly,yang2015large,gebru2017fine},  dogs~\cite{KhoslaYaoJayadevaprakashFeiFei_FGVC2011,parkhi12a,liu2012dog}, fashion~\cite{jia2020fashionpedia}, plants~\cite{nilsback2006visual,nilsback2008automated,kumar2012leafsnap}, 
food~\cite{bossard14,hou2017vegfru}, and the natural world~\cite{gvanhorn2018inat,van2021benchmarking}, to name a few. 

Datasets featuring images of different species of birds have been particularly popular in the vision community~\cite{wah2011caltech,berg2014birdsnap,van2015building,krause2016unreasonable}.
As a taxonomic group they present an interesting set of challenges that make them well suited for benchmarking advances in vision.   
For example, their appearance can differ based on life stage or sex, their shape can vary significantly, and some species can be very challenging for even expert humans to tell apart. 
Inspired by this, we propose a new multi-modal bird dataset that contains data from three different modalities: images, audio, and video.

\noindent{\bf Fine-Grained Video Datasets.} The most commonly used video action recognition datasets also tend to focus on coarse-grained concepts~\cite{kuehne2011hmdb,soomro2012ucf101,karpathy2014large,carreira2017quo,kay2017kinetics,damen2018scaling}, with some emphasizing temporal reasoning~\cite{goyal2017something,monfort2019moments,sevilla2021only}. Fewer fine-grained datasets exist, but those that do cover concepts such as sports~\cite{li2018diving48,piergiovanni2018fine,shao2020finegym} and cars~\cite{zhu2018fine,alsahafi2019carvideos}. 
Most relevant to this work are the small number of existing video datasets containing birds~\cite{ge2016exploiting,saito2016ibc127,zhu2018fine}, see Table~\ref{tab:existing_datasets} for an overview.
IBC127~\cite{saito2016ibc127} contains 8,014 videos across 127 bird categories. %
In the paper, experiments are performed for bird (127 classes) and action (4 classes) classification from video. %
VB100~\cite{ge2016exploiting} contains 1,416 videos from 100 bird species and evaluates on the task of species classification from video. 
While a small number of audio files are also available, no experiments are actually performed using this data. %
Finally, YouTube-Birds~\cite{zhu2018fine} contains 18,350 videos spanning the same 200 classes represented in the CUB200 image dataset~\cite{wah2011caltech}. 
The exact same set of videos are also used in~\cite{he2019new}. 
Experiments are performed on the task of bird classification from video, and they show that their approach gives a minor performance improvement compared to simple baselines\cite{wang2018temporal} which do not use any temporal information.   
The YouTube-Birds data is provided as a list of  YouTube video URLs, and at the time of writing only 17,031 videos are still publicly available. %

While these existing fine-grained datasets are very related to our work, they stop short of performing any cross-modal experiments, and do not show the benefit of audiovisual fusion methods for fine-grained categorization. Further, the distribution of data in these datasets is highly skewed and the included species were obviously dictated by data availability from web scraping. The SSW60 dataset provides a nearly uniform data distribution for a set geo-spatially co-located species. See the supplementary material for additional details.

\noindent{\bf Fine-Grained Audio Datasets.} There are numerous examples of human speech focused~\cite{garofolo1993timit,robinson1995wsjcamo}, coarse-grained audio classification~\cite{salamon2014dataset,piczak2015esc,fonseca2020fsd50k,gemmeke2017audio}, and binary sound event~\cite{mac2018bat,stowell2019automatic} datasets. 
However, in contrast to images, there are fewer established datasets for fine-grained audio classification.  
One task that is highly representative of a fine-grained audio challenge is that of species identification. 
As a result, there exists a number of audio datasets focused on species identification.  
Examples include bird~\cite{lostanlen2018birdvox,morfi2019nips4bplus,he2019new,cramer2020chirping,chronister2021annotated} and bat~\cite{zamora2016acoustic,roemer2021automatic} species classification. 
Like their image counterparts, these datasets can be challenging to collect and accurately annotate~\cite{baker2019deafening}. 
These annotation issues can also be compounded by factors such as background noise and low quality recordings. 
The audio recordings in the SSW60 dataset have been manually vetted by domain experts to ensure that the labels are reliable. %

\noindent{\bf Audiovisual Datasets.} In addition to visual content, video data can also contain rich and descriptive audio information. 
For some fine-grained concepts, this information can be highly complementary to the visual cues, see Fig.~\ref{fig:audiovisual_challenges}. 
Inspired by these types of relationships, the vision community has developed several benchmarks to facilitate the exploration of multi-modal reasoning. 
Several different approaches have been used to construct these types of datasets. 

The most basic approach is to query video media websites with keywords of interest, with the assumption that relevant sound events will also be present. This is the approach taken by the Flickr-SoundNet dataset~\cite{aytar2016soundnet}, which contains 2M video clips with audio downloaded from Flickr, and was queried using tags from YFCC100M~\cite{thomee2016yfcc100m}. 
An alternative approach is to use automatic filtering, \eg by making use of image or audio classification models. 
VGG-Sound~\cite{chen2020vggsound} consists of 200k, ten-second video clips from 300 different audio classes. The object that emits each sound is visible in the video clip, however, each clip is only labeled with one class even though multiple audio-visual events can be present. 
ACAV100M~\cite{lee2021acav100m} contains 100M ten-second clips and was constructed using an automatic curation pipeline that maximized the mutual information between the audio and visual channels. 
The final dataset construction approach is to manually annotate some or all of the data. 
Kinetics-Sounds~\cite{arandjelovic2017look} features 19k, ten-second, audio-visual clips covering 34 human orientated action classes. 
The videos are a subset of the Kinetics dataset~\cite{kay2017kinetics}, which were manually filtered to ensure the presence of the actions of interest. 
More detailed annotations include localizing sound events in time or space. 
AVE (Audio-Visual Event)~\cite{tian2018audio} is a subset of the AudioSet dataset~\cite{gemmeke2017audio}, and contains 4,143 ten-second videos covering 28 event categories with manually labeled temporal event boundaries. Each video contains at least one two-second long audio-visual event.
The LLP dataset~\cite{tian2020unified} contains 11,849 YouTube video clips with 25 event categories labeled. The goal of the dataset is audio-visual parsing, \ie deciding whether an event is audible, visible, or both. Manual temporal event annotations are provided for a subset of the videos. 
Finally, \cite{chen2021localizing} adds image bounding boxes to audible sound sources for 5k videos in VGG-Sound~\cite{chen2020vggsound}. In addition, the community has been working on audiovisual datasets for violence detection\cite{wu2020not}, as well as VQA~\cite{yun2021pano,li2022learning}

None of the above datasets explore the problem of fine-grained audiovisual reasoning. 
In this work, we make use of high quality image and audio classifiers in order to select video clips that are highly likely to contain the discriminative audiovisual events for a set of 60 bird species. 

\begin{table}[t]
\caption{Overview of existing bird datasets.
$^\circ$Only 17,031 videos are currently available online.
$^\star$Contains the same images as~\cite{wah2011caltech}.  
$^\dagger$Contains the same videos as~\cite{zhu2018fine}.
$^{\ddag}$Only spectrogram images are available, no audio files are included.
}
\footnotesize
\centering
\resizebox{0.6\linewidth}{!}{
\begin{tabular}{|l|llll|} \hline
dataset        & classes & images & videos & audio \\ \hline
CUB200~\cite{wah2011caltech}           &    200     &  11,788     &    -  &  -     \\
NABirds~\cite{van2015building}      &    555     &  48,562     &  -    &  -     \\
VB100~\cite{ge2016exploiting}       &    100     &  -    &  1,416     &   502      \\ 
IBC127~\cite{saito2016ibc127}      &    127         &  -         &  8,014    &  -     \\
YouTube-Birds~\cite{zhu2018fine}    &    200     &  11,788$^\star$    &     18,350$^\circ$  &  -     \\
PKU FG-XMedia~\cite{he2019new}      &    200     &  11,788$^\star$    &     18,350$^\dagger$  &  12,000$^{\ddag}$     \\ \hline
{\bf Ours}                         &    60 &  31,221      &     5,400  &    3,861    \\ \hline
\end{tabular}
}
\label{tab:existing_datasets}
\end{table}

\subsection{Multi-modal Learning}

\noindent{\bf Audiovisual Fusion.} %
There is large and growing literature on multi-modal fusion for audiovisual understanding. Early methods adopted straightforward early or score fusion strategies, \eg ~\cite{chen1998audio}. 
Subsequent research applied modality-specific networks with learning-driven information combinations in mid or late stage fusion strategies. 
Representative methods include activation summations~\cite{kazakos2019epic}, lateral connections~\cite{xiao2020audiovisual}, attention based re-weighting~\cite{fayek2020large}, among others. 
A comprehensive review can be found in~\cite{bayoudh2021survey}. 
The recent success of adopting transformer architectures in the vision~\cite{dosovitskiy2021an} and audio~\cite{gong21b_interspeech} communities empowered more advanced audiovisual fusion methods. %
A representative state-of-the-art work~\cite{nagrani2021attention} carefully studied audiovisual fusion with transformers and we adopt this as our primary baseline.

\noindent{\bf Cross-Modal Analysis.} \cite{kalogeiton2016analysing} defined and measured the impact of several different domain shift factors in the context of training object detectors on video frames and images. 
These factors included the accuracy of the training bounding boxes, appearance diversity, image quality, and object size.  
They showed that these factors, in combination, are almost completely responsible for the performance difference as compared to training and testing on the same domain. 
Their conclusion was that if one wants to achieve the best performance they should train and test on the same domain.  
In this work, we analyse domain differences arising from depictions of the same concept (\ie fine-grained bird categories) across different modalities.

\begin{figure*}[t]
\includegraphics[width=\linewidth]{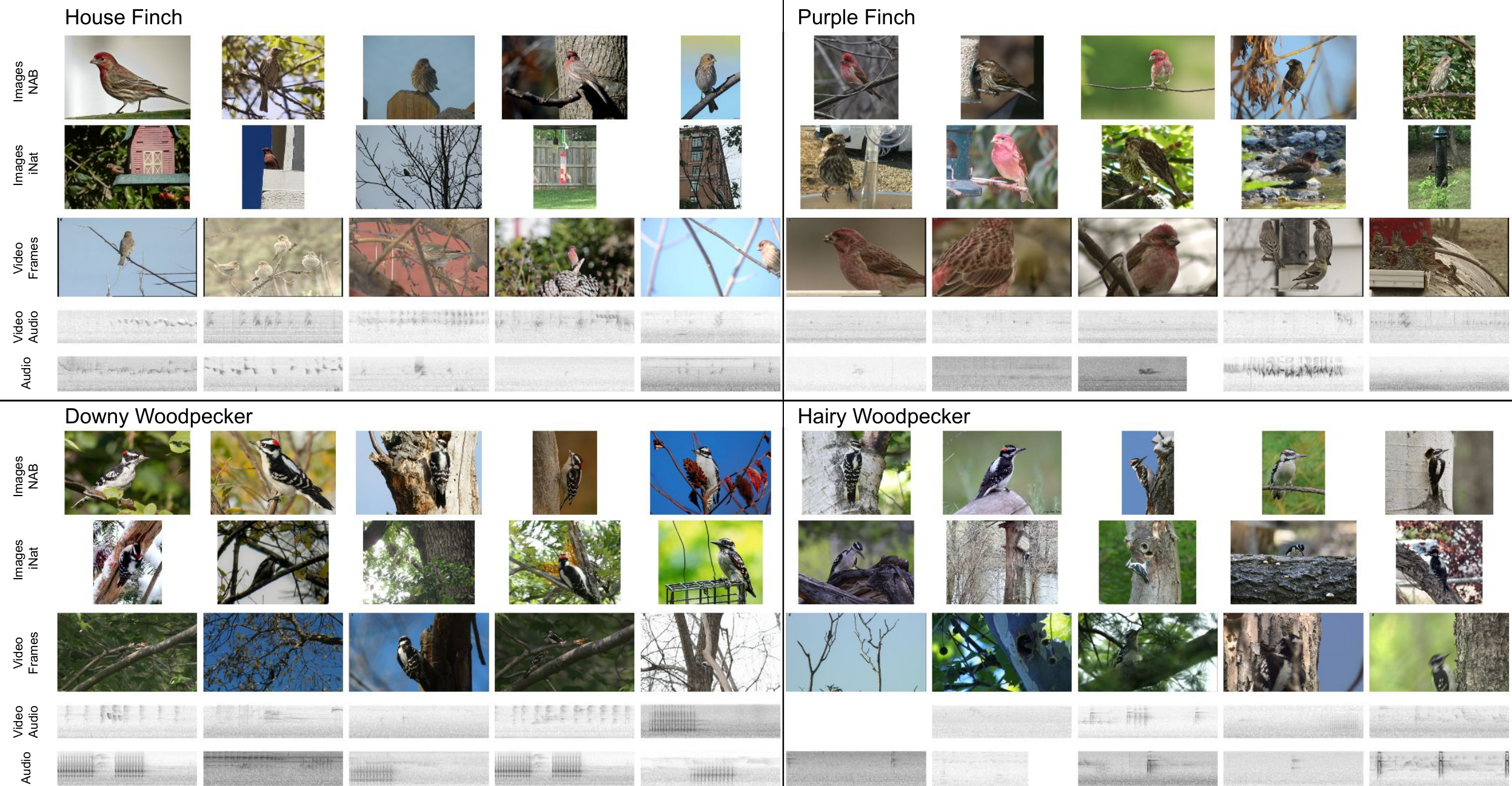}
\caption{Visual and audio examples (frames and spectrograms) for example bird species from the SSW60 dataset. 
Clockwise from top left: House Finch, Purple Finch, Hairy Woodpecker, and Downy Woodpecker. 
For each species, the five rows show modality samples from: (1) ``Images NAB'' - from NABirds~\cite{van2015building}; (2) ``Images iNat'' - from iNaturalist2021~\cite{van2021benchmarking}; (3) ``Video Frames'' - center frames; (4) ``Video Audio'' - spectrogram that covers the three seconds of audio near the center frame; and (5) ``Audio'' - spectrogram generated from three seconds near the center of the file. 
}
\label{fig:dataset_examples}
\end{figure*}

\section{SSW60 Dataset}
\label{sec:dataset}
In this section we describe the SSW60 dataset and the steps taken to construct it. 
The dataset is built around 60 species of birds that have a high propensity to be seen or heard on a live ``feeder-cam'' (\ie, a static camera monitoring a bird feeder) that is continuously recording in Ithaca, New York. %
These species, therefore, represent a realistic fine-grained challenge experienced by humans (unlike, \eg, CUB200 where the categories of birds come from all over the world). 
A model that can recognize these species and interpret their behaviors will be particularly relevant in assisting biologists with analyzing large collections of video footage from these cameras. 
We plan for future versions of this dataset to directly incorporate video from the live cams. 
For each of the 60 species we sampled data from three different modalities: videos (containing paired frames and audio), audio recordings, and images.
Here videos are unique as they contain both visual and audio modalities, while the additional unpaired audio recordings and image datasets only consist of one modality respectively. 
See Fig.~\ref{fig:dataset_examples} for examples from the various modalities and Table~\ref{tab:dataset_summary} for per-modality statistics.

\textbf{Video.}
The videos in SSW60 come from recordings archived at the Macaulay Library at the Cornell Lab of Ornithology~\cite{mlLibWeb}. 
These videos are contributed by professional and enthusiast videographers from the around the world, and can range in duration from a few seconds to multiple minutes. 
The camera view points are not fixed and can move in order to track the bird as it moves through the environment. 
Each video is associated with a particular ``target species'' that is known to be present in the video. 
For each video we isolated a ten-second clip where the task of species classification is particularly relevant. 
To accomplish this, we applied the following procedure: 1) For each of the 60 species of birds, we sampled all of their respective videos in the Macaulay Library. 
2) We then used an image based bird detector and classifier (trained on the 30M+ images from the Macaulay Library) to identify the sections of video where the target species was present. 
This gave us candidate video sections to extract ten-second clips. 
3) To further refine the candidate clips, we ran the Merlin Sound ID model~\cite{merlinSoundIdWeb}, a high performing acoustic bird classification model, across the audio tracks of the candidate clips to determine if the target species was vocalizing. 
4) For each video, we keep the clip with the highest likelihood of the target species vocalizing. 
5) Finally, for each species, we select 90 video clips 
for the dataset, where each clip comes from a unique video.

We found that most videos in the Macaulay Library do not have complete metadata indicating the exact recording time and date. %
We therefore split the video files into train and test sets by splitting on the videographers. 
All of the videos from a particular videographer are either in the train split or the test split. 
We found that this tactic was necessary to prevent multiple, highly similar videos uploaded by the same videographer winding up in both the train and test sets (a problem found in existing datasets~\cite{ge2016exploiting,saito2016ibc127}). 
All videos are converted to a frame rate of 25FPS. 
This modality is referred to as ``Video Frames'' in the experiment section when considering only the frames of the videos, and is referred to as ``Video Audio'' when considering only the audio channel. 

Note that some of the ten-second clips in SSW60 do not have the target species vocalizing; we do not treat this as a problem but view it as a challenge and inherent property of ``in-the-wild'' video. The process of a human uploading a video (as opposed to an audio recording) to the Macaulay Library, means that the videos in SSW60 will be biased towards visually relevant information for classification, as opposed to aurally relevant information. Using an acoustic classifier to find those sections of video with both visual and aurally relevant information helps mitigate this, but does not completely remove the visual bias.

\begin{table}[t]
\caption{
Summary of the train/test split sizes for each modality in SSW60, along with information about the number of examples per class.
}
\centering
\resizebox{0.7\linewidth}{!}{
\begin{tabular}{|l|l||l|l|l|l|}
\hline
              & source & total            & min          & max            &  median       \\ \hline
Images NAB  & \cite{van2015building} &  $5050$, $5171$  & $30$, $31$   &  $221$, $214$  &  $60$, $60$   \\
Images iNat & \cite{van2021benchmarking} &  $18000$, $3000$ & $300$, $50$  &  $300$, $50$   &  $300$, $50$  \\
\hline
Audio         & ours &  $2597$, $1264$  & $28$, $12$   &  $52$, $30$    &  $45$, $21$   \\
Video         & ours &  $3462$, $1938$  & $38$, $22$   & $68$, $52$     &  $59$, $31$   \\ 
\hline
\end{tabular}
}
\label{tab:dataset_summary}
\end{table}

\textbf{Audio.} 
All 60 bird species in SSW60 have unpaired audio recordings from the Macaulay Library. %
These recordings are unpaired in the sense that they do not have any associated visual data, \ie no videos or images. 
Each audio recording is annotated with a particular ``target species'' that is known to be vocalizing in the file. 
However, it is not specified at what moment in time the target species is vocalizing, and recordings can be multiple minutes long. 
We sampled audio recordings for each of our 60 species and had an expert ornithologist provide temporal onset and offset annotations for the target species. We then trimmed the audio files to ten-second clips that contain the target species' vocalization. The result is an expert curated audio dataset for each of the 60 species in SSW60. The audio files are stored in WAV format at a sampling rate of 22.05kHz.

Audio is split between train and test sets by ensuring that audio files from the same recording session are placed in the same split. %
This prevents models from exploiting common background noise that might be heard across multiple recordings from the same location and time. 
This modality is referred to as ``Audio'' in the experiment section.

\textbf{Images.} 
Finally, we also preform experiments with images from two existing datasets: NABirds~\cite{van2015building} and iNat2021~\cite{van2021benchmarking}. 
The 60 species in SSW60 conveniently overlap with the species in these existing datasets, and we incorporate all images available into SSW60 while maintaining the original train/test splits. 
For the NABirds dataset, we merged the respective ``visual categories'' that comprise each species. The images in NABirds are of particularly high quality, representing a best case scenario for visual classification (\ie someone using high quality camera equipment to carefully compose a photograph for the goal of visual identification). 
The images in iNat2021 are more mixed in terms of quality and therefore represent a more difficult visual classification task. See Fig.~\ref{fig:dataset_examples} for sample images from both datasets. 
These modalities are referred to as ``Images NAB'' and ``Images iNat'' in the experiment section.

\section{Methods}

We are interested in exploring fine-grained categorization in two areas: {\bf cross-modal analysis} and {\bf audiovisual fusion}.

For {\bf cross-modal analysis}, we assume a fixed backbone architecture that can be utilized for processing data from multiple modalities. For the audio modality, we convert the waveforms to spectrogram images. For videos, we adopt TSN~\cite{wang2018temporal} style methods using 2D image backbones to encode features and perform fusion on top of them. 
Our experimental procedure is straightforward: we train the backbone model using a particular training modality (see Sec~\ref{sec:dataset} for the options) and then evaluate the performance on an evaluation modality directly. As we have the same species in each modality, the trained backbone can be used directly on the evaluation modality. However, there is a domain transfer problem to consider (\ie moving from images to video frames), so we also evaluate the trained backbone by first fine-tuning the weights using the training split of the evaluation modality, and then evaluate on that evaluation modality. We use top-1 accuracy as the evaluation criteria for all experiments. Unlike existing bird video datasets~\cite{ge2016exploiting,saito2016ibc127,zhu2018fine}, our evaluation splits are uniform for each species, which makes top-1 accuracy across examples an unbiased assessment of performance. All backbone models are trained using softmax cross-entropy. 

In addition to cross-modality analysis, we study fine-grained {\bf audiovisual fusion} using the paired ``Video Frames'' and ``Video Audio'' data in SSW60. 
We adopt a transformer-based backbone and experiment with mid-fusion through the state-of-the-art multimodal bottleneck fusion approach of~\cite{nagrani2021attention}, as well as late and score-fusion. %
Thanks to the image and audio recordings provided in SSW60, we are able to study the effect of different pretraining dataset choices (\eg ImageNet, Images iNat, and  Images NAB) on audiovisual fusion.

\subsection{Implementation Details}
\label{implemenation_details}

\noindent{\bf Image Modality.} We adopt the standard ImageNet~\cite{russakovsky2015imagenet} training paradigm. During training, we randomly crop and resize a square portion of the image to 224$\times$224 pixels, followed by a random flip augmentation; during evaluation, the shorter edge of the image is resized to 256 pixels first, and then a center crop of 224$\times$224 is extracted for classification. We perform evaluation using a CNN-based ResNet50~\cite{he2016deep} and transformer-based ViT-B~\cite{dosovitskiy2021an} for experiments on the image modality, as they are the most popular choices in the image recognition community. 
Both are initialized with ImageNet pretrained weights.  

\noindent{\bf Audio Modality.} For audio processing, we convert the audio waveforms into spectrogram images. Concretely, the raw audio signal is resampled to a rate of 16kHz. We then apply the short-time Fourier transform algorithm using a window size of 512 and a stride length of 128. The frequency values are then transformed using the ``mel-scale'' with 128 bins. Finally we convert the magnitude values to decibel units and normalize to generate the final spectrogram image. This image is duplicated three times to create the RGB input for the network. For a 10-second long audio clip, the shape of the generated spectrogram image is approximately 128$\times$1250, where 128 is the number of mel-scaled frequency bands and 1250 is the temporal span. During training, we utilize two augmentations to avoid overfitting: time cropping and frequency masking~\cite{park19e_interspeech}. For time cropping, we randomly sample a window of length 400 time bins (spanning all 128 frequency bands) from the original spectrogram image (400 time bins corresponds to approximately 3 seconds of audio). For frequency masking, we randomly mask out 15 consecutive frequency bands. During evaluation, we densely sample five windows of length 400 time bins (spanning all 128 frequency bins) from the original spectrogram images using a stride of 150. We average the logits across the 5 windows to use as the final prediction. Earlier work~\cite{hershey2017cnn} used VGG-style~\cite{simonyan2014very} backbones which we also compare to for completeness.  

\noindent{\bf Video Frame Modality.} We adopt the segment sampling strategy of TSN~\cite{wang2018temporal} where 
we first divide the video clip into eight uniform segments. During training we randomly sample one frame from each segment, while for evaluation we select the center frame of each segment. Each of the eight selected frames is passed through the 2D ResNet50 or ViT-B backbone for feature extraction using images of size 224$\times$224 pixels. We then average the eight feature vectors to generate the final feature representation for the video clip. 
This global feature is then passed through a fully connected layer to produce a vector of logits. We initially conducted experiments with video-specific 3D convolution networks with dense frame sampling using S3D~\cite{xie2018rethinking}. However we found that S3D (pretrained on Kinetics-400~\cite{kay2017kinetics}) performed worse than our TSN-style baselines (pretrained on ImageNet) for SSW60. Furthermore, using a 2D network like ResNet50 or ViT-B as the backbone provides the flexibility for easily studying feature transfer between video frames and images. We leave experimentation with more sophisticated video backbones for future work.

\noindent{\bf Audiovisual Fusion.}
We briefly recap the transformer architecture and then describe how we conduct audiovisual fusion experiments. 
Given an input image or audio spectrogram, it is first divided into non-overlapping patches. Each patch is projected to a token using a linear layer and a special learnable classification token is added. More details can be find in the original ViT paper~\cite{dosovitskiy2021an}. After tokenization, the tokens are passed through a stack of transformer layers. 
We denote the input of the $l$-th layer as $\mathbf{z}^l$, which results in $\mathbf{z}^{l + 1} = trans\_layer_l(\mathbf{z}^{l})$. The computation inside $trans\_layer_l$ can be written as
\begin{align}
    \mathbf{y}^{l} &= \mathrm{MSA}(\mathrm{LN}(\mathbf{z}^l)) + \mathbf{z}^{l}, \label{eq:encoder_self_attention} \\
    \mathbf{z}^{l + 1} &= \mathrm{MLP}(\mathrm{LN}(\mathbf{y}^l)) + \mathbf{y}^l,
\end{align}
where LN denotes layer normalization, MSA denotes multi-head self-attention. 
In our audiovisual fusion, we use two identical $L=12$ layer transformers to take the visual and audio input separately. The forward process for the visual modality is thus $\mathbf{z}_{v}^{l + 1} = v\_trans\_layer_l(\mathbf{z}_{v}^{l})$, and  $\mathbf{z}_{a}^{l + 1} = a\_trans\_layer_l(\mathbf{z}_{a}^{l})$ for the audio modality. 

For mid-fusion, we use the state-of-the-art multimodal bottleneck transformer ~\cite{nagrani2021attention}. Here a set of learnable tokens $\mathbf{z}_{b}$ are used as the fusion bottleneck
\begin{align}
[\mathbf{z}_{v}^{l + 1}||\mathbf{\hat{z}}_{b}] & = v\_trans\_layer_l([\mathbf{z}_{v}^{l}||\mathbf{z}_{b}]), \\
[\mathbf{z}_{a}^{l + 1}||\mathbf{z}_{b}] & = a\_trans\_layer_l([\mathbf{z}_{a}^{l}||\mathbf{\hat{z}}_{b}]).
\end{align} 
$[.|| .]$ denotes the concatenation of tokens. 
In each layer, $\mathbf{z}_{b}$ first interacts with the visual tokens $\mathbf{z}_{v}^{l}$ and gets updated to $\mathbf{\hat{z}}_{b}$. Then $\mathbf{\hat{z}}_{b}$ interacts with the audio tokens $\mathbf{z}_{a}^{l}$ to finish the audio visual fusion. Following~\cite{nagrani2021attention}, we conduct this fusion in the last four layers of the transformer.  

For late fusion, we concatenate the class tokens of both modalities after the last transformer block, written as $[\mathbf{z}_{v}^{L+1}[0] || \mathbf{z}_{a}^{L+1}[0]]$ and apply a linear classifier on top of it. 
For score fusion, we take the predictions from both modalities and use a weighted sum of the combined final predictions. In practice, we use a weight of 0.5 for both the visual and audio modalities.

\section{Experiments}

We first perform cross-modal experiments on the video and audio modalities separately, and then explore multi-modal fusion for audiovisual categorization.

\textbf{Visual Modality Categorization.} Here we benchmark the performance achieved on the Video Frames of SSW60. Table~\ref{tab:cross_modal} (Left) shows the top-1 accuracy on the Video Frame test set when using a ResNet50 backbone trained on either the Images iNat, Images NAB, or the Video Frames training datasets. We split the results depending on whether we fine-tune (FT) the trained backbone on the Video Frames training dataset. Training directly on the Video Frames dataset achieves a top-1 accuracy of $54.92\%$. Interestingly, we see that evaluating the Images iNat model directly on the Video Frames achieves an even higher top-1 accuracy of $60.47\%$. This is further improved to $71.88\%$ when fine-tuning on the Video Frames train split. We compare these numbers to those achieved by a model trained on the Images NAB dataset: $24.05\%$ and $56.55\%$, top-1 accuracy respectively. The Images iNat dataset has more training samples than Images NAB, however, the images in the NABirds dataset are aesthetically higher quality (see Section~\ref{sec:dataset}). These results seem to indicate that performance on ``in-the-wild'' videos benefits more from ``lower quality'' training images. 

\begin{table}[]
\caption{
Top-1 accuracy on SSW60 Video-Frames using a ResNet50 backbone (left) and SSW60 Video-Audio using a ResNet18 backbone (right) when training on different datasets (columns). 
Results are presented with and without finetuning (FT) on the respective video modality. 
}
\centering
\resizebox{0.9\linewidth}{!}{
\begin{tabular}{|lccc|l|lcc|}
  \cline{1-4} \cline{6-8}
  \multicolumn{4}{|c|}{Cross-Modal - Video Frames}                                                                  &  & \multicolumn{3}{c|}{Cross-Modal - Video Audio}                                      \\ \cline{1-4} \cline{6-8} 
  \multicolumn{1}{|l||}{FT}     & \multicolumn{1}{l|}{Images iNat}  & \multicolumn{1}{l|}{Images NAB} & Video Frames &  & \multicolumn{1}{l||}{FT}     & \multicolumn{1}{l|}{Unpair Audio}     & Video Audio  \\ \cline{1-4} \cline{6-8} 
  \multicolumn{1}{|l||}{}       & \multicolumn{1}{l|}{60.47}        & \multicolumn{1}{l|}{24.05}     &      54.92   &  & \multicolumn{1}{l||}{}       & \multicolumn{1}{l|}{24.41}            &      10.37   \\ \cline{1-4} \cline{6-8} 
  \multicolumn{1}{|l||}{\cmark} & \multicolumn{1}{l|}{71.88}        & \multicolumn{1}{l|}{56.55}     &       -      &  & \multicolumn{1}{l||}{\cmark} & \multicolumn{1}{l|}{15.33}            &      -       \\ \cline{1-4} \cline{6-8} 
\end{tabular}
}
\label{tab:cross_modal}
\end{table}

\begin{table}[] %
\caption{ Comparison of audio backbones trained and tested on the unpaired audio modality in SSW60. 
All models are initialized from ImageNet pretrained weights. `$\uparrow$384' indicates that a model is fine-tuned on ImageNet with a higher resolution~\cite{touvron2021training} and `AS' is further fine-trained on AudioSet~\cite{gong21b_interspeech} before use. 
}
\centering
\resizebox{1.0\linewidth}{!}{
\begin{tabular}{ |l||c|c|c|c|c|c|c| } 
 \hline
 Backbone & VGG16 & VGG19 & ResNet18 & ResNet50 & ViT-B & ViT-B$\uparrow$384 & ViT-B$\uparrow$384 AS \\ 
 \hline
 Top 1 Acc & 52.1\% & 56.1\% & 59.01\% & 63.7\% & 66.8\% & 65.9\% & 67.4\% \\ 
 \hline
\end{tabular}
}
\label{tab:audio_backbones}
\end{table}

\textbf{Audio Modality Categorization.} We next benchmark the new unpaired audio dataset component of SSW60. Table~\ref{tab:audio_backbones} contains the results of these experiments. We trained and evaluated VGG16 and 19~\cite{simonyan2014very}, ResNet18 and 50~\cite{he2016deep}, and the transformer-based ViT-B~\cite{dosovitskiy2021an} architectures. As expected, we see a progression of top-1 accuracy as we move from older architectures ($52.1\%$ for VGG16) to the latest architectures ($66.8\%$ for ViT-B). We attempted to push accuracy further by using a ViT-B model pretrained on a higher resolution image input (224 vs 384), but we actually see performance decrease to $65.9\%$. However, if we take this higher resolution model and add an additional pretraining step of training on AudioSet~\cite{gemmeke2017audio} then we achieve a top-1 accuracy of $67.4\%$. 

We now benchmark the  Video Audio component of SSW60. %
For these experiments we chose a ResNet18 backbone for convenience, but expect a more powerful backbone to be slightly more performant (see Table~\ref{tab:audio_backbones} and Table~\ref{tab:av_fusion} (Direct Eval)).  
The obvious result is the low performance achieved when training exclusively with the Video Audio data, achieving a top-1 accuracy of just $10.37\%$. Directly using a model trained on the unpaired audio achieves $24.41\%$, a significant improvement. Interestingly, fine-tuning the unpaired audio model on the Video Audio training samples leads to a decrease in performance, down to $15.33\%$. This points to a recurring theme: video is biased to visual features (simply by the nature through which it was collected), and while it contains an audio channel, the ability to use the audio channel for classification appears to be difficult. We show in the next section however that it is possible to improve overall classification accuracy by incorporating audio.

\begin{table}[] %
\caption{
Audiovisual fine-grained categorization results on SSW60 videos using ViT models. 
We split results into two different scenarios: evaluation on the modalities individually (``Direct Eval'' and ``No Fusion'') and on both modalities together (``Fusion''). 
All numbers reflect top-1 accuracy.
``Direct Eval'' means we can conduct direct evaluation for modalities pretrained on datasets with the same 60 species.
``No Fusion'' means we take a pretrained network and fine-tune it on the respective modality from the SSW60 training videos. 
For the Mid and Late fusion algorithms in ``Fusion'', we initialize the model with pretrained weights from individual models trained on the ``Pretrain'' datasets. 
For Score fusion, we take the best individual model for each modality (considering both ``Direct Eval'' and ``No Fusion'' variants) and fuse their scores by a weighted sum. 
}
\centering
\resizebox{1.0\linewidth}{!}{
\begin{tabular}{|cc||cc|cc|ccc|}
    \hline
    \multicolumn{2}{|c||}{Pretrain} & \multicolumn{2}{c|}{Direct Eval}  &  \multicolumn{2}{c|}{No Fusion} & \multicolumn{3}{c|}{Fusion}\\ \hline
    Visual & Audio & Vid Frames & Vid Audio & Vid Frames & Vid Audio & Mid & Late & Score  \\
    \hline  \hline 
    ImageNet & ImageNet & - & - & 59.0\% & 14.3\% & 54.3\% & \textbf{59.8}\% & 58.9\%\\
    ImageNet & Unpair Audio & - & 28.3\% & 59.0\% & 30.4\% & 62.0\% & 62.5\% & \textbf{63.5}\% \\
    Images NAB & Unpair Audio & 60.0\% & 28.3\% & 64.4\% & 30.4\%& 67.5\% & \textbf{68.4}\% & 68.2\%\\
    Images iNat & Unpair Audio & 78.0\% & 28.3\%& 76.2\% & 30.4\%& 73.5\% & 78.3\% & \textbf{80.6}\% \\
    \hline
\end{tabular}
}
\label{tab:av_fusion}
\end{table}

\begin{figure*}[] %
    \begin{subfigure}[c]{0.3\textwidth}
    \includegraphics[width=\textwidth]{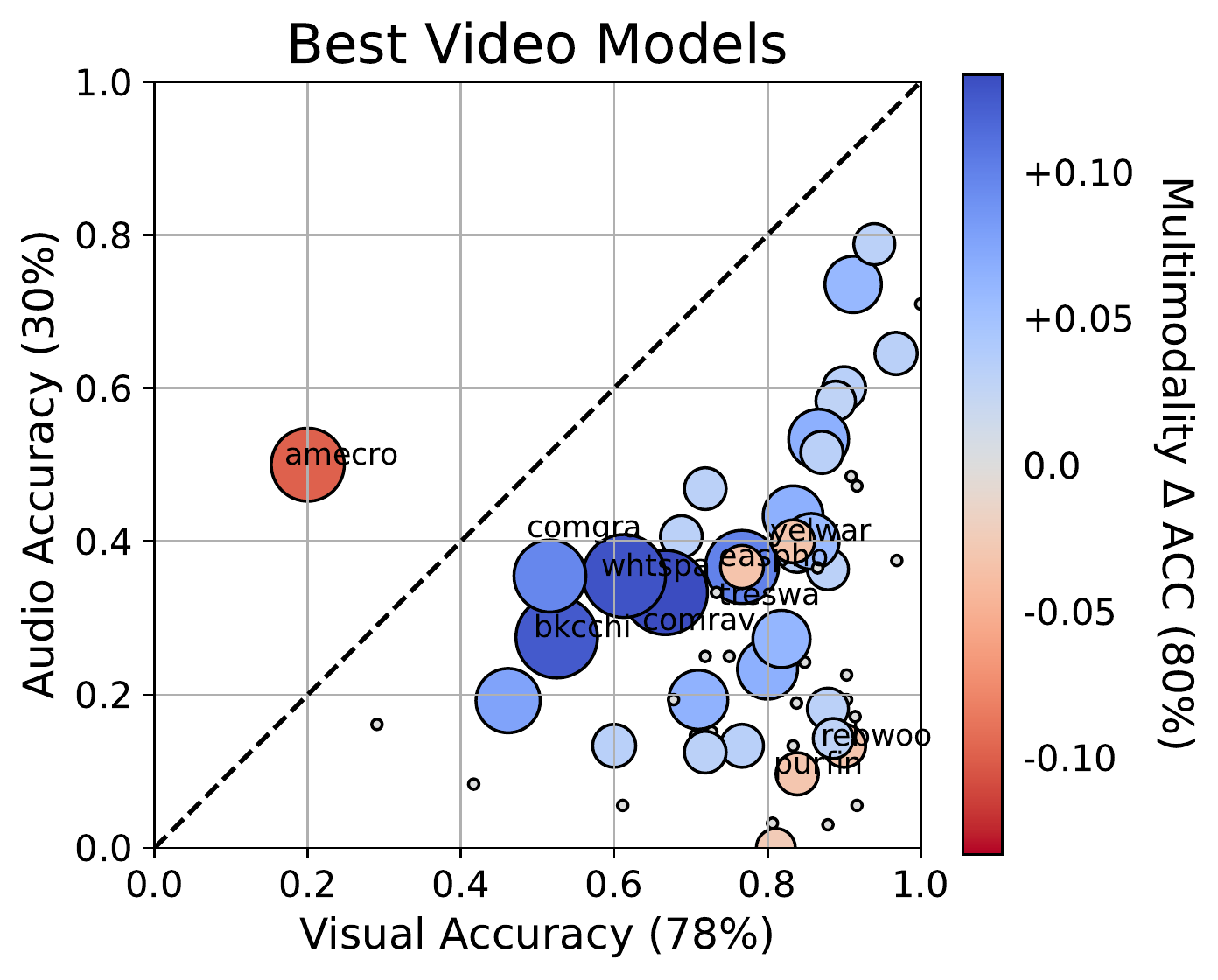}
  \end{subfigure}
  \begin{subfigure}[c]{0.7\textwidth}
    \includegraphics[width=\textwidth]{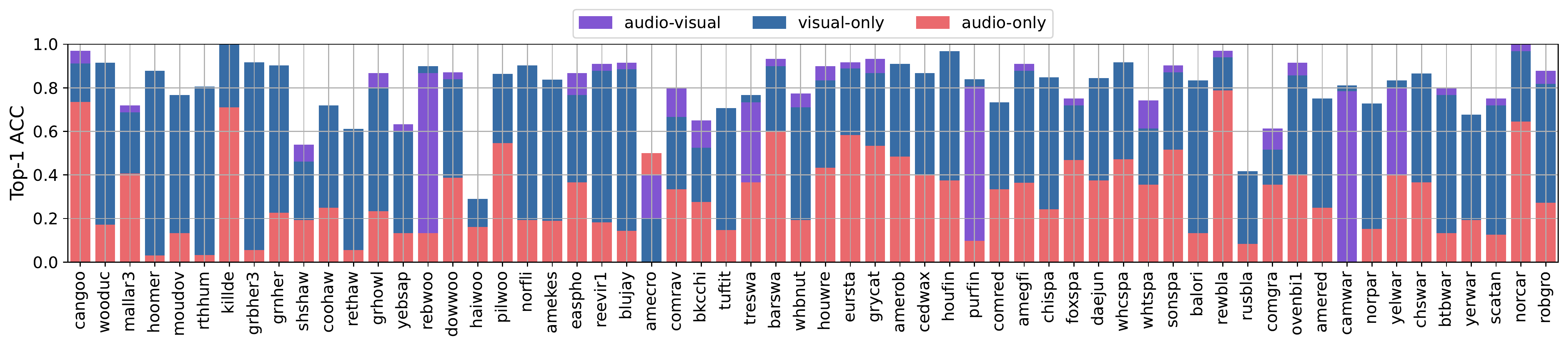}
  \end{subfigure}
  \caption{ 
  Per-species audio and visual modality performance along with the resulting audiovisual performance after score fusion.
  These results correspond to the bottom right fusion model in Table~\ref{tab:av_fusion}.
  The size and color of the dots on the scatter plot indicate the resulting top-1 accuracy change (from the best uni-modal model) when fusing the predictions for audiovisual classification. 
  Large blue dots correspond to better audiovisual accuracy. 
  Large red dots correspond to worse audiovisual accuracy.
  Species with the largest positive and negative audiovisual changes have been labeled.
  The bars for each species in the bar plot are ordered by the modality performance. 
  Purple bars on top reveal those species with improved audiovisual accuracy. 27 species improved, 27 species remained the same, and 6 species decreased after fusion.
  }
  \label{fig:av_non_video}
\end{figure*}

\textbf{Audiovisual Fine-Grained Categorization.} In contrast to the rich literature of audiovisual fusion on coarse-grained video datasets, audiovisual fine-grained categorization (FGC) remains under-explored due to a lack of appropriate datasets. 
SSW60 fills this gap and allows us to conduct a comprehensive analysis that explores the impacts of various pretraining and fusion methods on audiovisual FGC. 
We follow the paradigm employed by Nagragni \etal~\cite{nagrani2021attention} and use two uni-modal models to process the audio and visual modalities separately. 
We adopt the ViT-B~\cite{dosovitskiy2021an} backbone for both modalities (see Section~\ref{implemenation_details}). 
We are interested in two research questions: 1) What is the effect of different fusion methods? and 2) What is the effect of different pretraining datasets? 
For fusion methods, we use the state-of-the-art MBT~\cite{nagrani2021attention} as the mid-fusion algorithm, and compare to late and score fusion techniques. 
For pretraining datasets, we utilize ImageNet, Images NAB, and Images iNat for the visual modality, and ImageNet and unpaired audio recordings for the audio modality.
By construction, once a backbone has been trained on Images NAB, Images iNat, or the unpaired audio, we are able to directly evaluate on the corresponding modality of the SSW60 video dataset, since all datasets share the same 60 species.
Our results are summarized in Table~\ref{tab:av_fusion}. We also provide per-class analysis between uni-modal and audiovisual fusion performance in Fig.~\ref{fig:av_non_video}. Our best result on SSW60 (80.6\% top-1 accuracy) comes from fusing the scores of a visual model pretrained on Images iNat (and \textbf{not} fine-tuned on the SSW60 video frames), and an audio model pretrained on the unpaired audio and further fine-tuned on the audio channels of the SSW60 videos. 

We highlight three conclusions from our audiovisual fusion investigations. \textbf{First, the best result from audiovisual fusion is always better than training on each modality separately.} 
For each row in Table~\ref{tab:av_fusion}, the highest top-1 accuracy is always in the Fusion column, meaning that combining information from both modalities is always better than using a single modality.
This finding aligns well with our motivation of audiovisual fusion in Fig.~\ref{fig:audiovisual_challenges}. 
\textbf{Second, there is no ``best'' fusion method}. 
In the four different pretraining configurations, we find that late fusion works best half the time, and score fusion works best in the other half. 
It is interesting that the state-of-the-art mid-fusion method does not work as well as the simpler methods. 
We leave this as an open question for the community to explore more advanced mid-fusion methods for audiovisual FGC. 
\textbf{Third, pretraining on external datasets can be \textit{very} beneficial}.
We observe a $\sim20\%$ increase in top-1 performance when fine-tuning the ImageNet backbones on Images iNat and the unpaired audio (Fusion column, row 1 vs row 4 in  Table~\ref{tab:av_fusion}).

\section{Conclusion}

We present SSW60, a new dataset for advancing fine-grained audiovisual categorization.
This expert curated dataset provides researchers with the tools to explore categorization across three different modalities, enabling a comprehensive exploration of cross-modal and audiovisual fusion.   
Similar to how the CUB200 dataset paved the way for the larger, and better curated, NABirds and iNaturalist datasets, we envision SSW60 as a vital first step towards studying audiovisual fine-grained categorization. 
The availability of live ``feeder-cam'' video featuring the bird species in SSW60 also provides an interesting avenue for studying the deployment of trained models for real-time audiovisual categorization - an important problem for biodiversity monitoring.
At its current size, SSW60 can also be used as an evaluation dataset for self-supervised audiovisual models. %
We envision SSW60 broadly benefiting the vision community by providing ample directions for future work on FGC and video analysis more generally.

\noindent{\bf Limitations.} 
The size of the SSW60 dataset is a potential limitation for training models from scratch, which is why we used ImageNet pretrained models. ImageNet does contain $\sim$60 classes of birds, but all models started from an ImageNet pretrained backbone. The video and audio annotations in SSW60 are ``weak'' in the sense that they apply to the entire ten-second clip, as opposed to temporally localized annotations.

\noindent
\textbf{Acknowledgements.}
Serge Belongie is supported in part by the Pioneer Centre for AI, DNRF grant number P1. 
These investigations would not be possible without the help of the passionate birding community contributing their knowledge and data to the Macaulay Library; thank you!

\bibliographystyle{splncs04}
\bibliography{main}

\clearpage
\appendix

\setcounter{table}{0}
\renewcommand{\thetable}{A\arabic{table}}
\setcounter{figure}{0}
\renewcommand{\thefigure}{A\arabic{figure}}

\section{Existing Bird Video Datasets}

\begin{figure*}[ht]
\centering
\begin{subfigure}[b]{0.45\textwidth}
\includegraphics[width=\textwidth]{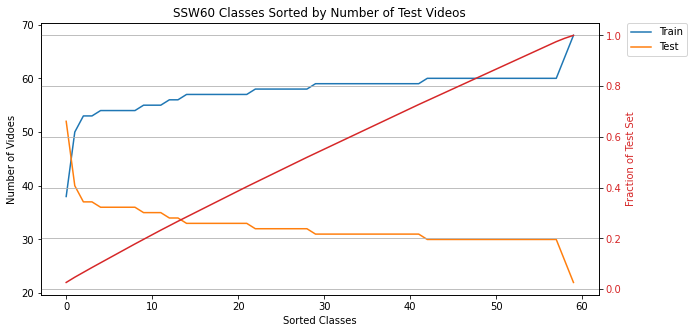}
\caption{SSW60}
\label{fig:ssw60_distribution}
\end{subfigure}
\begin{subfigure}[b]{0.45\textwidth}
\includegraphics[width=\textwidth]{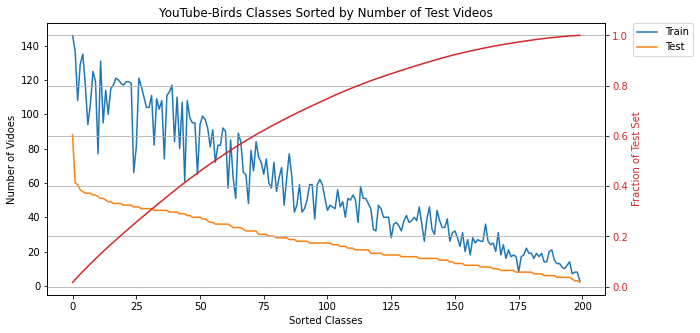}
\caption{YouTube-Birds}
\label{fig:ytb_distribution}
\end{subfigure}
\begin{subfigure}[b]{0.45\textwidth}
\includegraphics[width=\textwidth]{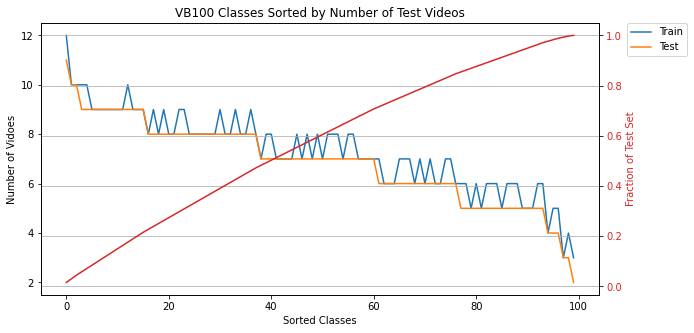}
\caption{VB100}
\label{fig:vb100_distribution}
\end{subfigure}
\begin{subfigure}[b]{0.45\textwidth}
\includegraphics[width=\textwidth]{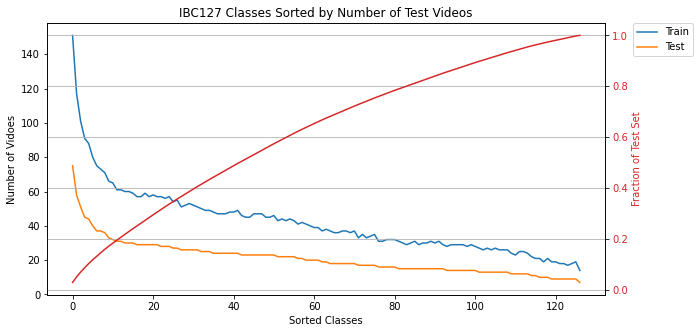}
\caption{IBC127}
\label{fig:ibc127_distribution}
\end{subfigure}
\caption{Train and test examples per species for various datasets. Note the (nearly) uniform train and test distributions for the SSW60 dataset compared to the other datasets.}
\label{fig:vid_dataset_distributions}
\end{figure*}

In this section we dive deeper into the existing bird video datasets~\cite{ge2016exploiting,saito2016ibc127,zhu2018fine} and discuss why they were not suitable for our investigations. See Table~\ref{tab:existing_dataset_stats} and Table~\ref{tab:existing_dataset_train_test_stats} for overview statistics comparing the different datasets. As mentioned in the main paper, \emph{none} of these prior works explore cross modality or audiovisual fine-grained categorization. For reference when comparing the datasets, the distribution of train/test videos in the SSW60 dataset can be seen in Fig.~\ref{fig:ssw60_distribution}.

\subsection{YouTube-Birds}

The YouTube-Birds dataset~\cite{zhu2018fine} is a collection of 18,350 videos that cover the same 200 categories as the CUB200 dataset~\cite{wah2011caltech}. The dataset is provided as a collection of YouTube links, with no information regarding which section of a video is relevant for classification (see Table~\ref{tab:existing_dataset_stats} for statistics on the video duration). At the time of writing only 17,031 videos are still available (a link attrition rate of 7\% over 3 years). The distribution of both the train and test videos per category is non-uniform, see Fig.~\ref{fig:ytb_distribution}. In the benchmark experiments for this dataset, it is unclear whether the authors used top-1 accuracy averaged across all test videos (``micro'') or if they first computed top-1 accuracy for each species and then averaged those values to get overall top-1 accuracy (``macro''). Given that the test distribution is non-uniform, ``micro'' accuracy would give a very skewed sense of performance, since the 56 categories with the most train data have over 50\% of the test videos. 

Unlike websites organized around a particular fine-grained domain (like iNaturalist~\cite{iNatWeb} or the Macaulay Library~\cite{mlLibWeb}), YouTube has no mechanisms to vouch for the reliability of tags or labels applied to videos (\ie to confirm if the species labelled as being present are actually correct). Therefore the creators of YouTube-Birds had to query for videos using CUB200 category names (presumably searching the titles and descriptions for text matching the names) and then ``used a crowd sourcing system to annotate the videos''~\cite{zhu2018fine}. No details are given describing the skill of the annotators, and it is well documented that crowd workers (\eg those on Amazon Mechanical Turk) can provide noisy labels when annotating fine-grained data~\cite{van2015building}. We therefore expect the error rate in YouTube birds to be at least has high as it is in the CUB200 dataset: 5\%~\cite{van2015building}. While conducting a thorough cleanup of YouTube-Birds is beyond the scope of this work, we did find particularly high error rates in those categories with few videos (\eg only 1 / 5 videos were relevant for the ``024.Red\_faced\_Cormorant" category, and 6 / 11 videos were relevant for the ``151.Black\_capped\_Vireo'' category).  

The lack of a well defined 10 second clip also makes YouTube-Birds unwieldy for the task of classification. While some videos focused on a single individual, in others, the birds played a small role. For example, which species should a model focus on in this video: \url{www.youtube.com/watch?v=wiCr5Yqo5y0} - which is assigned to the `151.Black\_capped\_Vireo'' category in the dataset? There are two different species, each in clear focus during different sections of the video, but neither are necessarily the focus of the video. In addition, large portions of the video consist of an interview with a human. The task is ambiguous for evaluation, and confusing for training. While narrowing a video down to a 10-second clip does not completely alleviate this problem, it does certainly help.  

We chose not to use the YouTube-Birds dataset due to the challenges associated with downloading (potentially broken) YouTube links, the high probability of labeling errors, and the issue of untrimmed video clips. One final inconvenience of the YouTube birds dataset is that while the authors matched the categories of the CUB200 dataset, they used a different label assignment for their annotations. While just an inconvenience, it highlights that this dataset poses serious obstacles for effective analysis of cross modal performance. 
Our SSW60 dataset aims to alleviate many of the issues listed above, \ie it will be distributed as a single download as opposed to a list of YouTube links, it has been curated by bird experts so the label quality is very high, and it contains 10-second video clips which focus on the bird of interest.

\begin{table*}[!htb]
\footnotesize
\caption{Video duration (in seconds) stats for existing bird video datasets.
$^\circ$18,350 videos originally. 
}
\centering
\begin{tabular}{|l|llllll|} \hline
dataset                             & classes & videos           & Avg Dur & Med Dur &  Min Dur & Max Dur  \\ \hline
VB100~\cite{ge2016exploiting}       &    100  &  1,416           & 32.6 & 32.14 & 4.60 & 200.83       \\ 
IBC127~\cite{saito2016ibc127}       &    127  &  8,014           &  31.2 & 28.66 & 3.00 & 266.72  \\
YouTube-Birds~\cite{zhu2018fine}    &    200  &  17,031$^\circ$  &  60.5 & 49.04 & 0.76 & 465.2     \\
 \hline
{\bf SSW60} (Ours)                          &    60 &  5,400  & 9.7 & 9.96 & 2.20 & 9.96    \\ \hline
\end{tabular}
\label{tab:existing_dataset_stats}
\end{table*}

\begin{table*}[!htb]
\caption{Train and test stats for existing bird video datasets, for each class. Means are rounded to the nearest tenth.
$^\circ$18,350 videos originally.
}
\footnotesize
\centering
\begin{tabular}{|l|lllllll|} \hline
dataset                             & classes & videos           & Total  & Avg &  Med & Min & Max  \\ \hline
VB100~\cite{ge2016exploiting}       &    100  &  1,416           &  730, 686 & 7.3, 6.9 & 7, 7 & 3, 2 & 12, 11      \\ 
IBC127~\cite{saito2016ibc127}       &    127  &  8,014           & 5343, 2671 & 42.1, 21.0 & 37, 19 & 14, 7 & 151, 75  \\
YouTube-Birds~\cite{zhu2018fine}    &    200  &  17,031$^\circ$  &  11735, 5296  & 58.7, 26.5 & 50, 25 & 3, 2  &  146, 88  \\
 \hline
{\bf SSW60} (Ours)                          &    60 &  5,400  & 3462, 1938 & 57.7, 32.3 & 59, 31 & 38, 22 & 68, 52      \\ \hline
\end{tabular}
\label{tab:existing_dataset_train_test_stats}
\end{table*}
 
\subsection{VB100}

The VB100 dataset~\cite{ge2016exploiting} is a collection of $1,416$ videos covering 100 bird species, with a non-uniform distribution of train and test images per species, see Fig.\ref{fig:vb100_distribution}. The authors do not provide information on the source of the videos, but upon visual inspection it is highly likely that most of these videos came from the Internet Bird Collection (IBC) website. The media on this website has since been incorporated into the Macaulay Library\footnote{www.macaulaylibrary.org/the-internet-bird-collection-the-macaulay-library/}. One challenge of using media from IBC is that one has to be careful with how videos are separated into train and test splits. Many videos from IBC are actually shorter clips from a longer recording session or part of a longer original video. For example the VB100 videos corresponding to ``American\_Rock\_Wren\_00001.mp4"\footnote{www.macaulaylibrary.org/asset/201760451} and ``American\_Rock\_Wren\_00002.mp4"\footnote{www.macaulaylibrary.org/asset/201760441} are from the same recording session, but one is a test video and the other is a train video in the VB100 dataset. This leaks information across the train/test splits, providing an opportunity for models to `cheat'. We aim to mitigate this from occurring in SSW60 by placing all the videos from a particular videographer into either the train or test split. 

We chose not to use the VB100 due to its small size, random collection of species (see the IBC127 discussion below), and problems with the existing train/test splits. Also it should be noted that there are other minor issues with the dataset, \eg the annotation files accompanying the dataset are incorrectly formatted, so that ``Sandwitch\_Tern'' in the annotations files corresponds to the ``Sandwich\_Tern'' directory of videos (note the typo). %

\subsection{IBC127}

The IBC127 dataset~\cite{saito2016ibc127} is a collection of $8,014$ videos covering 127 species of birds. The videos in this dataset were originally downloaded from the Internet Bird Collection (IBC) website. As mentioned above, the media on this website has since been incorporated into the Macaulay Library. Similar to VB100, the IBC videos must be split into train and test splits carefully, so as to prevent leakage of information. In the paper the authors state that they ``use 5,343 videos for learning and 2,671 videos for testing''~\cite{saito2016ibc127},  however these splits are not included with the dataset. It is unclear whether the authors attempted to maintain a uniform or non-uniform test set for each species. The dataset also does not provide user IDs for the videos, so we are unable to ensure that we create reliable train/test splits. We assume the authors used a non-uniform test split (because the numbers easily match those provided by the authors under this assumption), and generated the data in Table~\ref{tab:existing_dataset_train_test_stats} for the IBC127 dataset by randomly creating a 2/1 train/test split for each species (to match the authors' 5,343 / 2,671 split).

Overall, IBC127 is actually a reasonable dataset to start from. It has an imbalanced data problem, and the train/test conundrum is a serious problem, but we could have invested time manually (or automatically) to review the videos. However, a big problem with IBC127 is the random collection of bird species that comprise the dataset (a problem that affects the VB100 dataset as well). These species were clearly chosen because they satisfied some data quantity threshold when the authors were downloading videos. As we are interested in image and audio modalities, each of which would have their own data collection requirements, we wanted to avoid a `hodgepodge' of bird species. We built SSW60 around 60 species of birds that all occur in a specific geographic region. This makes the classification task realistic, and also means that progress on the dataset directly impacts the biologists working on these species. The live ``feeder-cams'' mentioned in the main body of the paper is a prime example of a real world use case for an audiovisual classifier built on SSW60. 

We chose not to use the VB100 dataset due to its missing metadata for train/test set creation, skewed video distribution, and its random collection of species.

\section{Visual Cross-Modality Results}
\label{sec:visual_cross_modality}

In Table~\ref{tab:viz_results} we provide detailed results for cross-modality experiments on the visual modalities of the SSW60 dataset. Results on rows 5, 8, 15, 18, and 22 are also presented in Table 3 of the main paper. For completeness, we present results for models that have either been pretrained on ImageNet or simply randomly initialized. These experiments also explore the linear classifier setting for both training and domain transfer evaluation settings. All datasets (regardless of source) use the same 60 categories. Each row is a different experiment. Simply put, each experiment consists of (1) choosing a training dataset, (2) training a model, (3) choosing an evaluation dataset, and (4) evaluating the trained model. These experiments explore various tactics for training the classifier and for handling the domain shift when shifting to different evaluation datasets. Columns:\\
\begin{itemize}
\itemsep-0.4cm
\item \textbf{Initialization:} specifies whether the ResNet-50 backbones starts from ImageNet weights or randomly intialized weights. \\
\item \textbf{Pretrain dataset:} specifies the source of training data used for the experiment. This is either the NABirds dataset (NAB), the iNaturalist dataset (iNat'21), or the frames of the videos from the SSW60 video clips. \\
\item \textbf{Pretrain modality:} specifies whether the ResNet-50 backbone was trained using images or video clips. See Section 4.1 in the main paper for details on how the different modalities are used to train the backbone. \\
\item \textbf{Pretrain method:} specifies how the ``Pretrain dataset'' was used to train the ResNet-50 backbone. Options are: \textbf{Linear}: we leave the ResNet-50 backbone weights fixed and we train a linear classifier by extracting a feature vector for each train sample in the ``Pretrain dataset''. \textbf{Finetune}: we fine-tune the weights of the ResNet-50 backbone using the ``Pretrain dataset.'' \\
\item \textbf{Evaluation dataset:} specifies the source of evaluation data for measuring top-1 accuracy. This is either the NABirds dataset (NAB), the iNaturalist dataset (iNat'21), or the frames of the videos from the SSW60 video clips. \\
\item \textbf{Evaluation modality:} specifies whether the trained model (either a linear classifier or a fine-tuned network) was evaluated using images or video clips. See Section 4.1 in the main paper for details on how the different modalities are used for evaluation. \\
\item \textbf{Evaluation method:} specifies how we used the ``Evaluation dataset'' to evaluate the trained model. Options are: \textbf{Direct}: we directly evaluate on the test samples of the evaluation dataset. \textbf{Linear:} we train a linear classifier using the \textit{training} samples from the evaluation dataset, and then evaluate on the test samples. \textbf{Finetune}: we fine-tune the weights of the ResNet-50 model on the \textit{training} samples from the evaluation dataset, and then evaluate on the test samples.
\end{itemize}

\begin{table*}[t]
\caption[foo bar]{Full results for \textbf{visual} cross-modality experiments. For all experiments we use a ResNet-50 backbone. See Sec.~\ref{sec:visual_cross_modality} for a description of the experiment setup and column explanations.
}
\centering
\resizebox{1.0\linewidth}{!}{
\begin{tabular}{l|c|c|c|c|c|c|c|c}
\footnotesize 
\# & Initialization & \begin{tabular}[c]{@{}c@{}}Pretrain\\ dataset\end{tabular} & \begin{tabular}[c]{@{}c@{}}Pretrain\\ modality\end{tabular} & \begin{tabular}[c]{@{}c@{}}Pretrain\\ method\end{tabular} & \begin{tabular}[c]{@{}c@{}}Evaluation \\ dataset\end{tabular} & \begin{tabular}[c]{@{}c@{}}Evaluation\\ modality\end{tabular} & \begin{tabular}[c]{@{}c@{}}Evaluation\\ method\end{tabular} & 
\begin{tabular}[c]{@{}c@{}}Top-1 acc.\\ (\%)\end{tabular} \\
\hline
1 & ImageNet & NAB & Image & Linear & NAB & Image & Direct & 79.20 \\
2 & ImageNet & NAB & Image & Finetune & NAB & Image & Direct & 90.31 \\
3 & Random & NAB & Image & Finetune & NAB & Image & Direct & 59.56 \\
\hline
4 & ImageNet & NAB & Image & Linear & SSW60 & Video & Direct & 17.44 \\
5 & ImageNet & NAB & Image & Finetune & SSW60 & Video & Direct & 24.05 \\
6 & Random & NAB & Image & Finetune & SSW60 & Video & Direct & 3.41 \\
7 & ImageNet & NAB & Image & Finetune & SSW60 & Video & Linear & 46.54 \\
8 & ImageNet & NAB & Image & Finetune & SSW60 & Video & Finetune & 56.55 \\
\hline
9 & ImageNet & iNat'21 & Image & Linear & NAB & Image & Direct & 75.94 \\
10 & ImageNet & iNat'21 & Image & Finetune & NAB & Image & Direct & 91.67 \\
\hline
11 & ImageNet & iNat'21 & Image & Linear & iNat'21 & Image & Direct & 53.40 \\
12 & ImageNet & iNat'21 & Image & Finetune & iNat'21 & Image & Direct & 75.20 \\
13  & Random & iNat'21 & Image & Finetune & iNat'21 & Image & Direct & 51.57 \\
\hline
14 & ImageNet & iNat'21 & Image & Linear & SSW60 & Video & Direct & 37.87 \\
15 & ImageNet & iNat'21 & Image & Finetune & SSW60 & Video & Direct & 60.47 \\
16 & Random & iNat'21 & Image & Finetune & SSW60 & Video & Direct & 24.36 \\
\hline
17 & ImageNet & iNat'21 & Image & Finetune & SSW60 & Video & Linear & 73.63 \\
18  & ImageNet & iNat'21 & Image & Finetune & SSW60 & Video & Finetune & 71.88 \\
19 & Random & iNat'21 & Image & Finetune & SSW60 & Video & Linear & 45.72 \\
20  & Random & iNat'21 & Image & Finetune & SSW60 & Video & Finetune & 46.44 \\
\hline
21  & ImageNet & SSW60 & Video & Linear & SSW60 & Video & Direct & 35.60 \\
22  & ImageNet & SSW60 & Video & Finetune & SSW60 & Video & Direct & 54.92 \\
23  & Random & SSW60 & Video & Finetune & SSW60 & Video & Direct & 10.06 \\
\hline
24  & ImageNet & SSW60 & Video & Linear & NAB & Image & Direct & 13.85 \\
25  & ImageNet & SSW60 & Video & Finetune & NAB & Image & Direct & 18.45 \\
26  &  Random & SSW60 & Video & Finetune & NAB & Image & Direct & 1.59 \\
\hline
27  & ImageNet & SSW60 & Video & Finetune & NAB & Image & Linear & 8.97 \\
28  &  ImageNet & SSW60 & Video & Finetune & NAB & Image & Finetune & 56.91 \\
29 &  Random & SSW60 & Video & Finetune & NAB & Image & Linear & 8.41 \\
30  & Random & SSW60 & Video & Finetune & NAB & Image & Finetune & 58.67 \\
\hline
\end{tabular}
}

\label{tab:viz_results}
\end{table*}

\section{Audio Augmentations}

We employ augmentations at training time for both the visual and audio modalities, see Section~4.1 in the main paper for descriptions. 
In Table~\ref{tab:audio_augs} we provide results when we disable different augmentation types on the audio modality. 
The model is equivalent to the ViT-B backbone results in Table~4 of the main paper. 
We can see that the addition of augmentations improves performance. 

\begin{table}[h!]
\caption{Audio augmentation ablations using a ViT-B backbone.}
\centering
\begin{tabular}{c|c|c}
\hline
No augmentation & + time crop & + frequency mask \\ \hline
44.1 & 60.6 & 66.8 \\ 
\end{tabular}
\label{tab:audio_augs}
\end{table}

\section{Video Clip Examples}
\begin{figure*}[t]
\includegraphics[width=\linewidth]{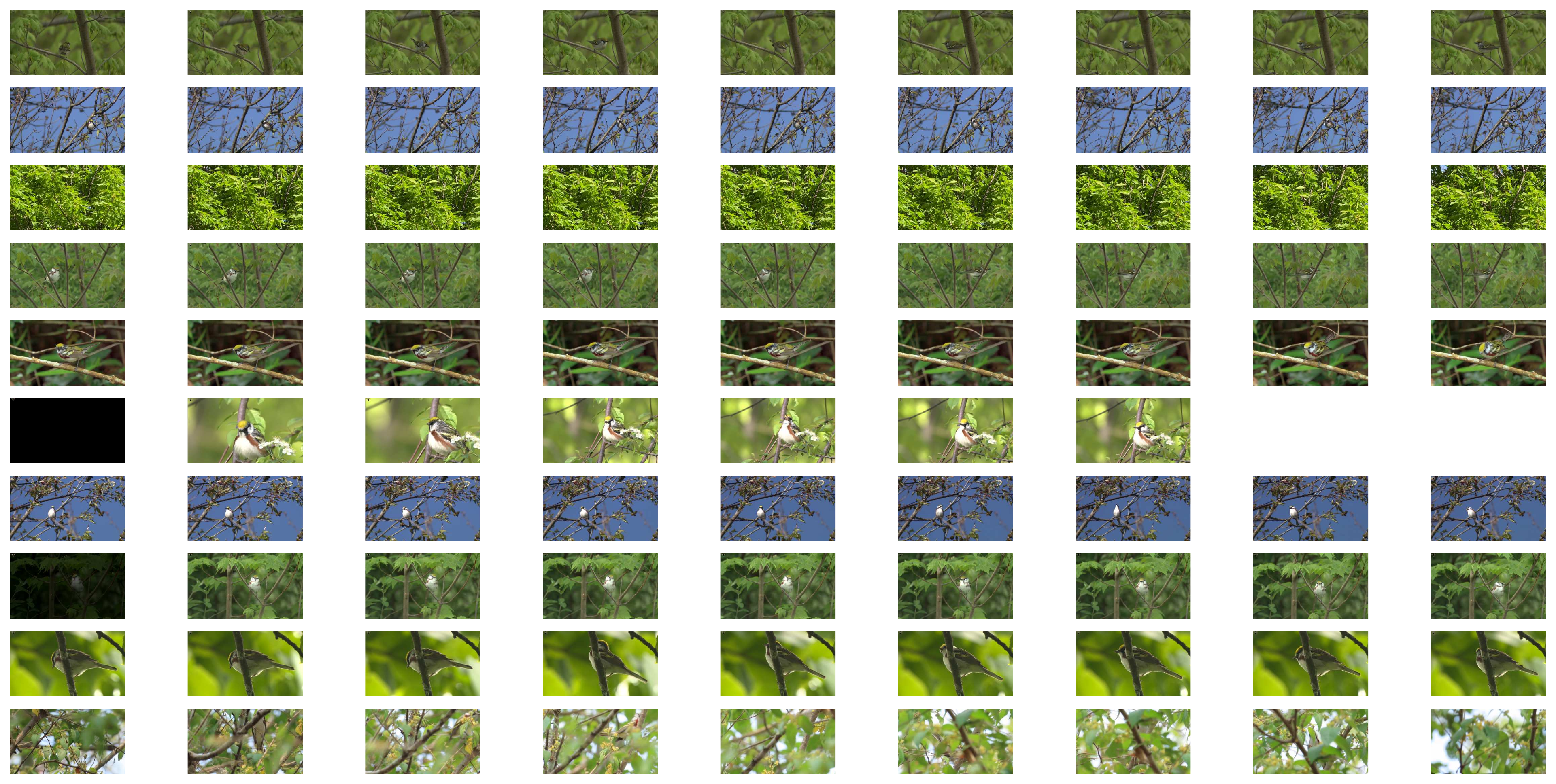}
\caption{1Hz frames from Chestnut-sided Warbler videos from our SSW60 dataset.}
\label{fig:1hz_ex1}
\end{figure*}

\begin{figure*}[t]
\includegraphics[width=\linewidth]{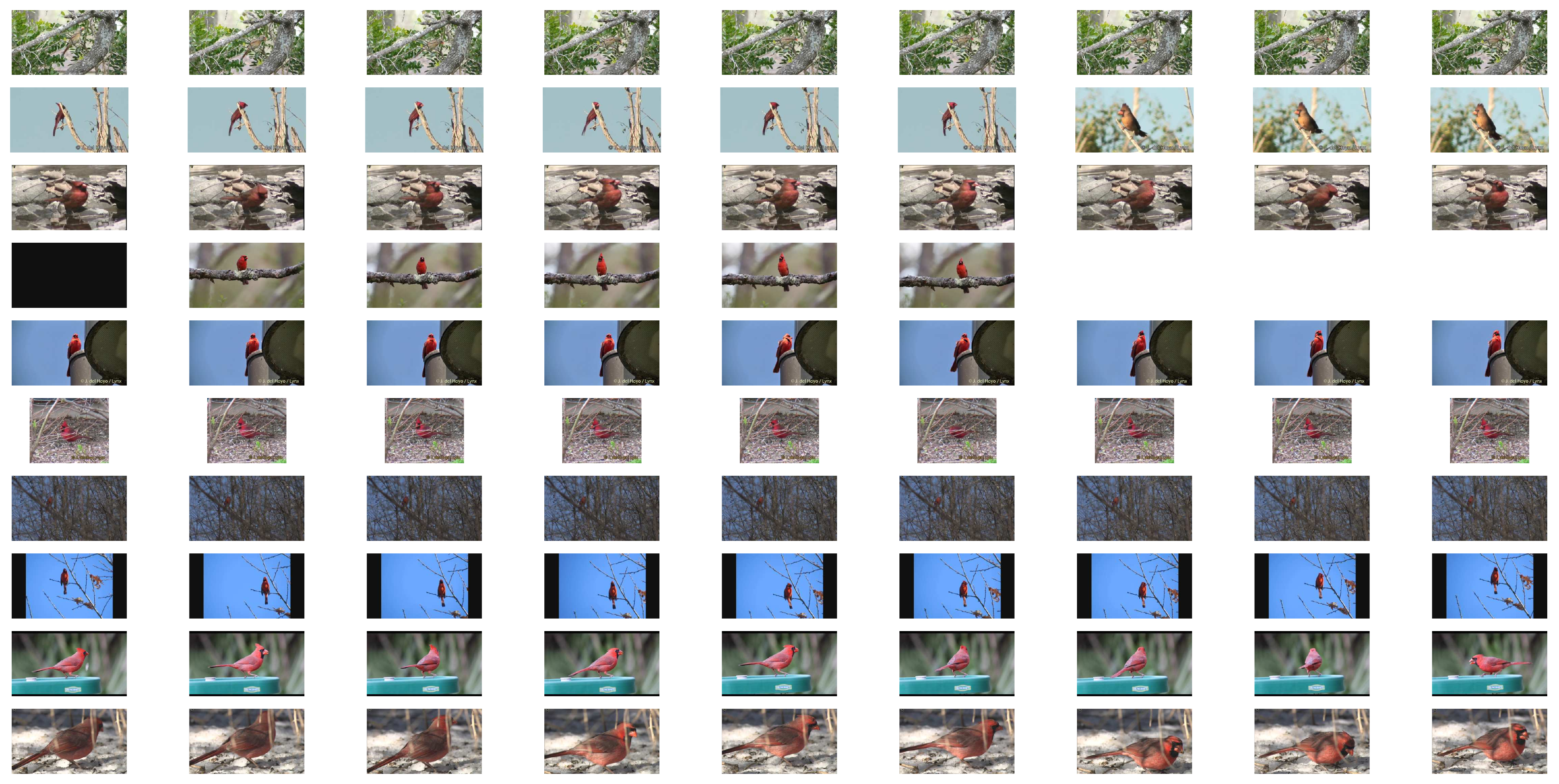}
\caption{1Hz frames from Northern Cardinal videos our SSW60 dataset.}
\label{fig:1hz_ex2}
\end{figure*}

\begin{figure*}[t]
\includegraphics[width=\linewidth]{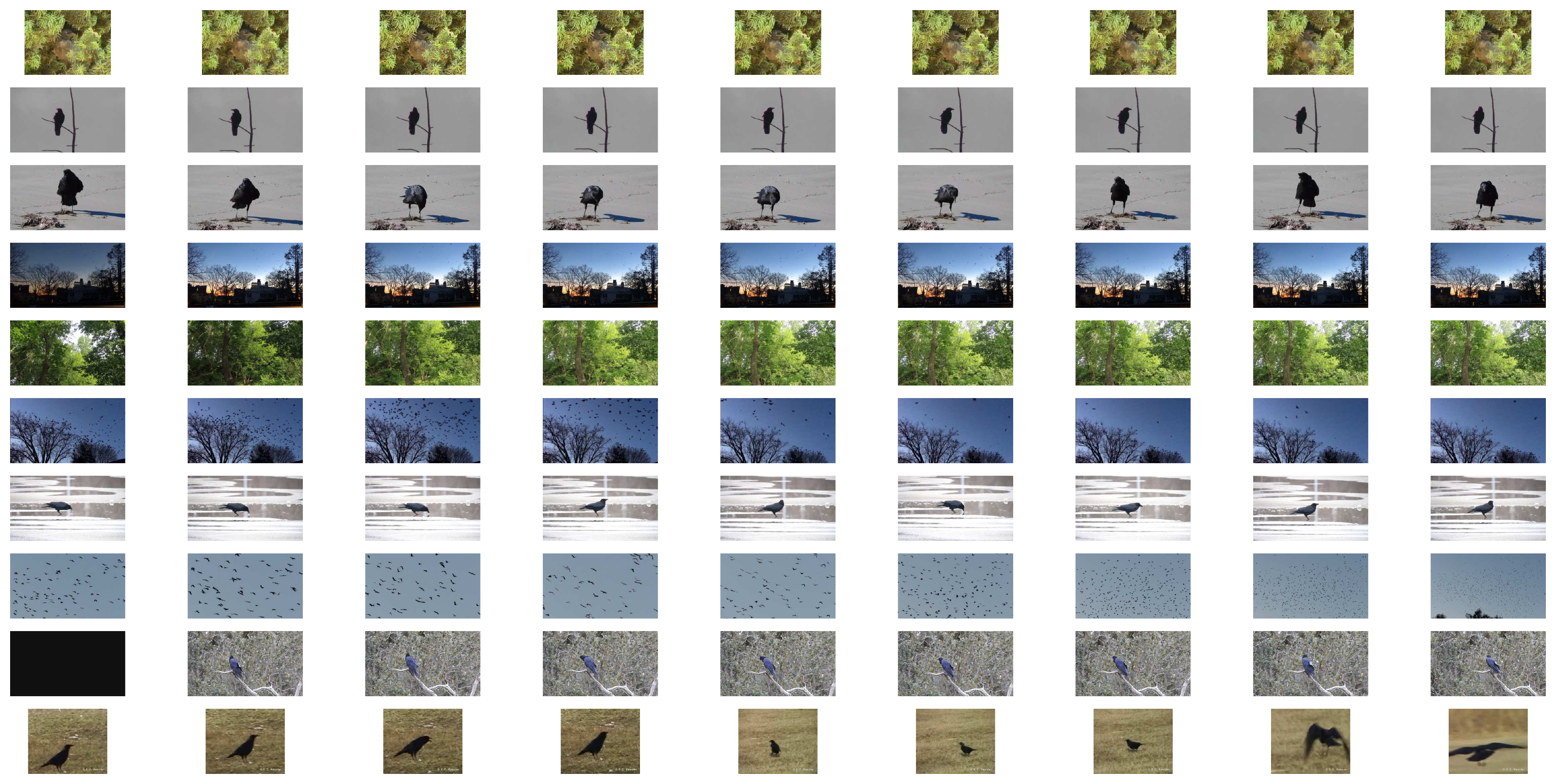}
\caption{1Hz frames from American Crow videos our SSW60 dataset.}
\label{fig:1hz_ex3}
\end{figure*}

\begin{figure*}[t]
\includegraphics[width=\linewidth]{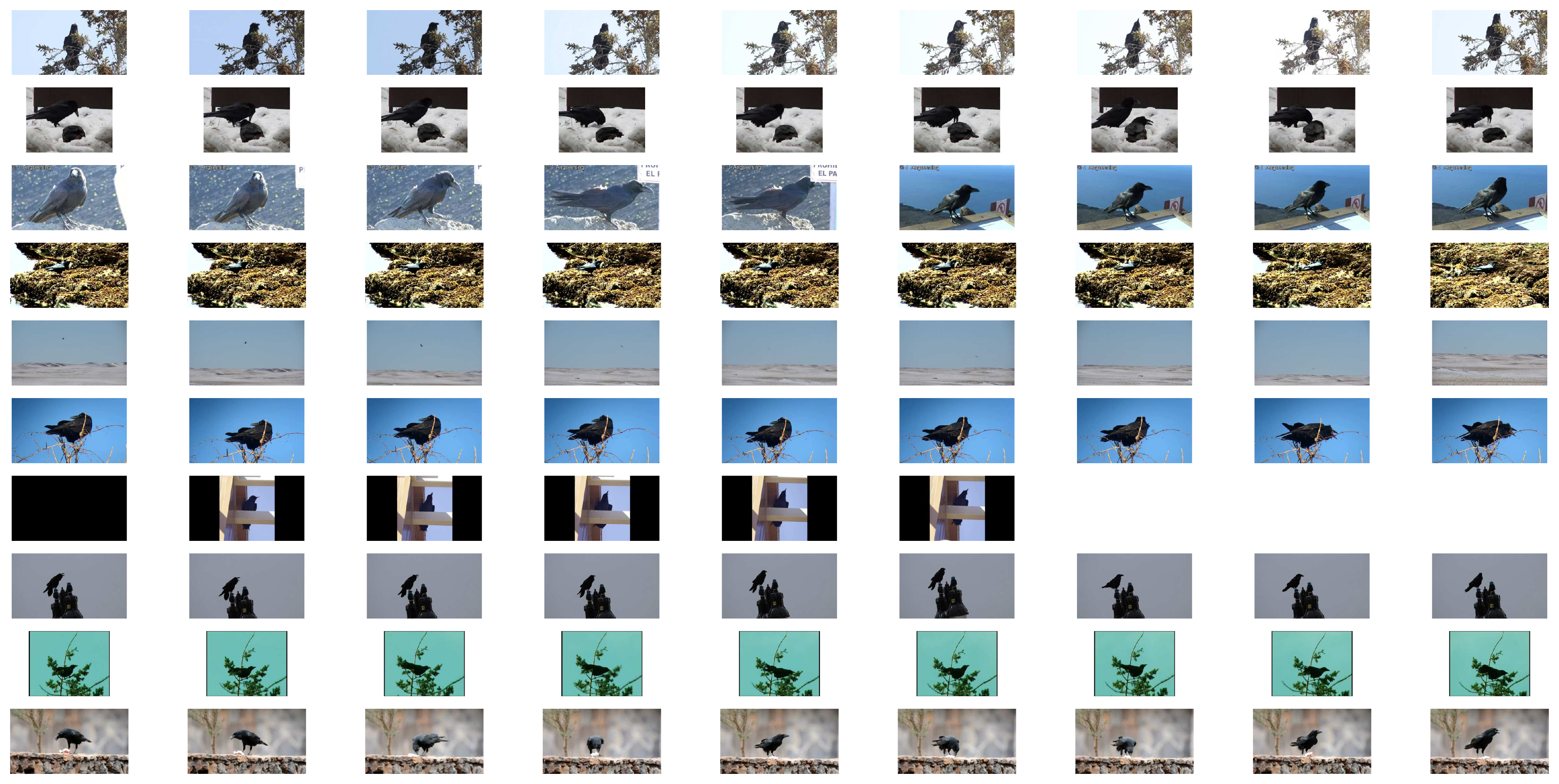}
\caption{1Hz frames from Common Raven videos our SSW60 dataset.}
\label{fig:1hz_ex4}
\end{figure*}

In Figs.~\ref{fig:1hz_ex1}, \ref{fig:1hz_ex2}, \ref{fig:1hz_ex3}, and \ref{fig:1hz_ex4} we show frames sampled at 1Hz from randomly sampled videos from our SSW60 dataset. 

\end{document}